# All You Need Is Synthetic Task Augmentation

Guillaume Godin[1]

[1]Osmo Labs PCB New York, USA

Corresponding authors: guillaume@osmo.ai

**Abstract**

- Injecting rule-based models like Random Forests[1] into differentiable neural network frameworks remains an open challenge in machine learning. Recent advancements have demonstrated that pretrained models can generate efficient molecular embeddings. However, these approaches often require extensive pretraining and additional techniques, such as incorporating posterior probabilities, to boost performance. In our study, we propose a novel strategy that jointly trains a single Graph Transformer[2] neural network on both sparse multitask molecular property experimental targets and synthetic targets derived from XGBoost[3] models trained on Osmordred[4] molecular descriptors. These synthetic tasks serve as independent auxiliary tasks. Our results show consistent and significant performance improvement across all 19 molecular property prediction tasks. For 16 out of 19 targets, the multitask Graph Transformer outperforms the XGBoost single-task learner. This demonstrates that synthetic task augmentation is an effective method for enhancing neural model performance in multitask molecular property prediction without the need for feature injection or pretraining.

  **Scientific Contribution** We demonstrate that Graph Transformer models can surpass multiple XGBoost baseline on physical molecular properties by leveraging a synergetic training approach that combines experimental and synthetic tasks augmentation, without relying on pretraining.

---

# Introduction

Integrating rule-based models like Random Forests into differentiable neural networks for molecular property prediction presents significant challenges. Recent research has investigated strategies to bridge the methodological gap between symbolic models and neural networks, aiming to combine the interpretability of rule-based approaches with the representational power of deep learning.

One of the pioneering methods in this domain is Transformer-CNN[5], which leverages random-to-random SMILES[6] translation in an encoder-decoder framework to extract embeddings, subsequently processed by a convolutional neural network for property prediction. This approach has shown competitive performance on a range of molecular property benchmarks and challenges[7].

Building upon this, TabPFN[8] framework introduces task-aware pretraining, utilizing large-scale synthetic data to generate effective molecular embeddings. By pretraining on synthetic tasks and fine-tuning on specific property prediction tasks, TabPFN has shown improved performance, particularly in low-data regimes. However, a direct comparison with Transformer-CNN is still lacking in the literature.

In the ChemProp[9] model, domain-specific molecular descriptors can be injected into the graph neural network to enrich its internal representations. This acknowledges that graph neural networks alone may overlook complex cheminformatics features. However, the benefit of this approach depends on how closely the injected features align with the prediction task. Poor alignment may introduce task-feature mismatch, limiting performance or leading to overfitting.

Prior work[10] suggests that multitask learning improves performance when tasks are correlated, as shared representations naturally promote inductive transfer. This is logical, as the model's final layer influences all tasks, making them interdependent by construction[11]. Each task's output becomes a linear combination of the final layer's output, reinforcing the interconnectedness of tasks within the model. Conversely, multitask learning can be ineffective or even detrimental when the target properties are weakly correlated.

---

[5] Pavel Karpov, Guillaume Godin, and Igor V. Tetko, "Transformer-CNN: Swiss Knife for QSAR Modeling and Interpretation," *Journal of Cheminformatics* 12, no. 1 (December 2020): 17, https://doi.org/10.1186/s13321-020-00423-w.

[6] David Weininger, "SMILES, a Chemical Language and Information System. 1. Introduction to Methodology and Encoding Rules," *Journal of Chemical Information and Computer Sciences* 28, no. 1 (February 1, 1988): 31–36, https://doi.org/10.1021/ci00057a005.

[7] Andrea Hunklinger et al., "The openOCHEM Consensus Model Is the Best-Performing Open-Source Predictive Model in the First EUOS/SLAS Joint Compound Solubility Challenge," *SLAS Discovery* 29, no. 2 (March 2024): 100144, https://doi.org/10.1016/j.slasd.2024.01.005.

[8] Noah Hollmann et al., "Accurate Predictions on Small Data with a Tabular Foundation Model," *Nature* 637, no. 8045 (January 9, 2025): 319–26, https://doi.org/10.1038/s41586-024-08328-6.

[9] Esther Heid et al., "Chemprop: A Machine Learning Package for Chemical Property Prediction," *Journal of Chemical Information and Modeling* 64, no. 1 (January 8, 2024): 9–17, https://doi.org/10.1021/acs.jcim.3c01250.

[10] Fabio Capela et al., "Multitask Learning On Graph Neural Networks Applied To Molecular Property Predictions" (arXiv, 2019), https://doi.org/10.48550/ARXIV.1910.13124.

To the best of our knowledge, no prior work has simultaneously trained on both experimental and synthetic targets within a unified multitask model. There were other methods proposed to leverage synthetic knowledge. One paper was found to train only on synthetic data from Nvidia[12], specifically to be pre-trained on synthetic data then fine-tuned on real data for images segmentation in two independent steps for a single target model. Another study[13], prior to TabPFN, employed synthetic data to shape the latent space of an autoencoder, in a way conceptually similar to the TabPFN model. Another approach was proposed to train at least one important physical feature from *ab initio* synthetic data and add them into the embedding of the Chemprop model[14]. Another idea was to use single task models knowledge distillation to guide the multitask model[15]. This knowledge distillation approach incorporates both synthetic and experimental data but merges them into a single output via a weighted objective, rather than modeling each separately. It was also mentioned to use masked knowledge in computer vision but this is not predicting two tasks but instead filtering the DOI (domain of Importance) for simplify the model task[16].

In this work, we propose Synthetic Task Augmentation: a direct training strategy that jointly learns from experimental and synthetic tasks, eliminating the need for feature injection, pretraining, or distillation.

Our main contributions are the following:

- We demonstrate that the inclusion of synergistic synthetic tasks is key to effective multitask learning.

---

[11] "A New Neural Kernel Regime: The Inductive Bias of Multi-Task Learning | OpenReview," accessed May 3, 2025, https://openreview.net/forum?id=APBq3KAmFa.

[12] Jonathan Tremblay et al., "Training Deep Networks with Synthetic Data: Bridging the Reality Gap by Domain Randomization" (arXiv, 2018), https://doi.org/10.48550/ARXIV.1804.06516.

[13] Nicolae C. Iovanac and Brett M. Savoie, "Improved Chemical Prediction from Scarce Data Sets via Latent Space Enrichment," *The Journal of Physical Chemistry A* 123, no. 19 (May 16, 2019): 4295–4302, https://doi.org/10.1021/acs.jpca.9b01398.

[14] Kevin P. Greenman, William H. Green, and Rafael Gómez-Bombarelli, "Multi-Fidelity Prediction of Molecular Optical Peaks with Deep Learning," *Chemical Science* 13, no. 4 (2022): 1152–62, https://doi.org/10.1039/D1SC05677H.

[15] Chaeyoung Moon and Dongsup Kim, "Prediction of Drug–Target Interactions through Multi-Task Learning," *Scientific Reports* 12, no. 1 (October 31, 2022): 18323, https://doi.org/10.1038/s41598-022-23203-y.

[16] "Pseudo Label-Guided Multi Task Learning for Scene Understanding | OpenReview," accessed May 4, 2025, https://openreview.net/forum?id=b4Phn_aTm_e.

- We propose a novel framework that integrates experimental and synthetic tasks within a single multitask graph neural network, eliminating the need for separate pretraining stages.

- Our method outperforms naive Graph Transformer on all molecular property prediction tasks, highlighting the efficacy of combining experimental data with synthetic task augmentation.

- We provide empirical evidence that this approach enhances model performance without relying on extensive feature engineering or pretraining, simplifying the model development process.

This study underscores the potential of integrating synthetic tasks directly into multitask learning frameworks, offering a more efficient and effective pathway for molecular property prediction.

# Data

A dataset of 19 targets was collected from various sources including OpenOCHEM[17]. Those targets are all related to molecular properties of interest for a total of 28987 unique molecules with at least one target per molecule (see Table 1). We use the RDkit[18] to canonize the Smiles and to merge the whole entries into a 28987 x 19 sparse matrix. All the models were trained using the exact same split seed into 5 groups of molecules to simplify the comparison between our models and repeat 5 times for 5 different seeds.

Table 1: List of molecular properties targets

| Target | Data Size | Unit |
|---|---|---|
| Density | 10305 | $kg/m_3$ |
| Refractive index | 9928 | None |
| **log Pow** | 5665 | log10 |
| **log Solubility*** | 10218 | log10 mol/L |
| **Abrahams (6)** | 9060 V<br>7680 L<br>8700 E<br>7894 B<br>8204 S | None |

---

[17] Tetko, Igor V., Martin Šícho, and Guillaume Godin, "Openochem/Openochem," Python, April 23, 2025, https://github.com/openochem/openochem.
[18] Greg Landrum et al., "Rdkit/Rdkit: 2023_09_1 (Q3 2023) Release Beta" (Zenodo, October 6, 2023), https://doi.org/10.5281/ZENODO.591637.

|  | 8706  A |  |
|---|---|---|
| Boiling Point | 11053 | °K |
| Melting Point | 7208 | °K |
| Hansens (3) | dP, dH, dD   9699 | None |
| Flashpoint | 7705 | °K |
| Vapor pressure | 3728 | log10 Pa |
| Volatility | 3728 | log10 ug/L |
| Henry Constant | 4173 | Unit less |

List of targets selected from several databases including OpenOCHEM and curated. In bold targets with highly correlated Osmordred features. * molecules can contain metal atoms.

For almost all targets (except for logSolubility see table 1), we only keep the organic molecules without metals. We remove all single heavy atom molecules like "Methane: C" for example as it is not compatible with graphs convolution neural networks. The same dataset was used for all models in this study.

We use the recently developed Osmordred collection of features (see table 2 and 3) as input for a XGBoost model. Osmordred is a new c++ of the RDkit that includes on top of padel and mordred features (see table 2) additional features (see table 3).

**Table 2 Padel/Mordred like 2D descriptors**

| ABCIndex | DetourMatrix | MolecularId |
|---|---|---|
| AcidBase | DistanceMatrix | PathCount |
| AdjacencyMatrix | EState | Polarizability |
| Aromatic | EccentricConnectivityIndex | RingCount |
| AtomCount | ExtendedTopochemicalAtom | RotatableBond |
| Autocorrelation | FragmentComplexity | **SLogP** |
| BCUT | Framework | TopoPSA |
| BalabanJ | HydrogenBond | TopologicalCharge |
| BaryszMatrix | InformationContent | TopologicalIndex |
| BertzCT | KappaShapeIndex | VdwVolumeABC |
| BondCount | Lipinski | VertexAdjacencyInformation |
| RNCGRPCG | **LogS** | WalkCount |
| CarbonTypes | **McGowanVolume** | Weight |
| Chi | MoeType | WienerIndex |
| Constitutional | MolecularDistanceEdge | ZagrebIndex |

List of 2D descriptors available in Osmordred generally part of Padel and Mordred, in bold features highly correlated to our targets.

More precisely, several issues in the computation were fixed and few methods were completely rewrite to improve the speed especially like for example InformationContent that now is based on a very fast algorithm inspired by the initial work of Basak[19] with a simplification added in order to optimize comparison of large keys of all atoms at a given radius. Now only the current radius key in an internal cluster is compared for memory improvement and sorted key comparisons.

An independent study of Osmodred features was proposed by Pat Walters[20] and demonstrated a good model performances on Polaris[21] difficult data using a "5 x 5 CV" method in combination with Lightboost[22] method instead of XGBoost.

**Table 3 Osmordred extension 2D descriptors**

---

[19] A. B. Roy et al., "NEIGHBORHOOD COMPLEXITIES AND SYMMETRY OF CHEMICAL GRAPHS AND THEIR BIOLOGICAL APPLICATIONS," in *Mathematical Modelling in Science and Technology*, ed. Xavier J. R. Avula et al. (Pergamon, 1984), 745–50, https://doi.org/10.1016/B978-0-08-030156-3.50138-7.
[20] "Practical_cheminformatics_posts/Adme_comparison/ADME_Comparison.Ipynb at Main · PatWalters/Practical_cheminformatics_posts," accessed May 3, 2025, https://github.com/PatWalters/practical_cheminformatics_posts/blob/main/adme_comparison/ADME_Comparison.ipynb.
[21] "Polaris," accessed May 3, 2025, https://polarishub.io.
[22] Robert P. Sheridan, Andy Liaw, and Matthew Tudor, "Light Gradient Boosting Machine as a Regression Method for Quantitative Structure-Activity Relationships" (arXiv, 2021), https://doi.org/10.48550/ARXIV.2105.08626.

| | | |
|---|---|---|
| Pol | AlphaKappaShapeIndex | Triplet indexes |
| MR | HEState | Frags |
| Flexibility | BEState | AddFeatures |
| Schultz | **Abrahams**[23] | |

List of 2D descriptors added to cover larger target space, in bold highly correlated to our tasks.

As XGBoost allows nan in inputs we did not impute Osmordred features. Finally, some features may have very large absolute values; a pre transformation was applied based on a threshold absolute value 33 to normalize extreme values. The pre transformation function is Arcsinh. We have decided to remove the unique values features of the dataset, the very low variance values and the very highly correlated duplicate columns reducing to 2449 features for training other the 3585 features initially available in Osmordred. In our method we use the 5 times 5 cross validation to the dataset.

# Results

We trained the "XGBoost model" independently as target by target using only Osmordred features input (see supplementary part for performances and CV method). Our final "XGBoost model" will be used for our Synthetic Task Augmentation.

This model will be our benchmark to train one naive graph transformer (GT) multitask model. We use torch, pytorch geometric to implement our GT model. After the training, we compare the single GT model with the "XGBoost model" and observe a similar performance in general (See Figure 1).

One important remark for clarifying extreme results in figure 1: $1-R^2$ (Var Explained) can be seen as a zoom with much more detail on the local effect than $R^2$ (Direct). Specifically, "V abraham" also known as "McGowen characteristic volume" is an exact value already calculated in Osmordred features (r2>0.9998), obvious XGBoost model is perfect for this target, while the Graph Transformer has a decrease of performance compared to the exact XGBoost solution (see figure 1). The $1-R^2$ (Var Explained) showed a very high modification moving from aka 0.9999 to 0.999. While metric $R^2$ (Direct) showed a very slightly reduction of the performance. We got an almost similar effect for another Abraham feature "E" that is also part of the Osmordred features and known to be also a very easy task with almost perfect prediction.

[23] James A. Platts et al., "Estimation of Molecular Linear Free Energy Relation Descriptors Using a Group Contribution Approach," *Journal of Chemical Information and Computer Sciences* 39, no. 5 (September 27, 1999): 835–45, https://doi.org/10.1021/ci980339t.

More precisely, "V Abraham" experimental data is not ideally correlated with "McGowen characteristic volume" in Osmordred. We have a very high probability that the toolkit to generate the McGowen Volume from the original (experimental) data does not have exactly the same aromaticity as the RDkit. A "little modification between chemical toolkits" implies a side effect on property prediction. The author has observed a similar effect even for Information Content values at radius 0. To avoid confusion, Information Content at radius 0 is just simply based on cluster count per atomic number (e.g. number of Carbons, number of Hydrogens, etc…). After carefully checking those deviations at radius 0, it was found only a modification of the number of hydrogen atoms of a molecule (+/-1) between the RDkit and the Basak toolkit. Which resulted in a difference of aromatic behaviour.

Our next step is to evaluate what is the impact on performance by adding the synthetic "XGBoost model" tasks as additional targets to the Graph Transformer model. We predict simultaneously now 38 targets instead of 19 before.

We got a 100% performance improvement with all our targets (see figure 2). This systematic improvement not only delivers a better model to GT without synthetic training tasks but also delivers a better model compared to the "XGBoost model" for 16 targets over 19.

Figure 1: comparison of 19 XGBoost versus our Graph Transformer

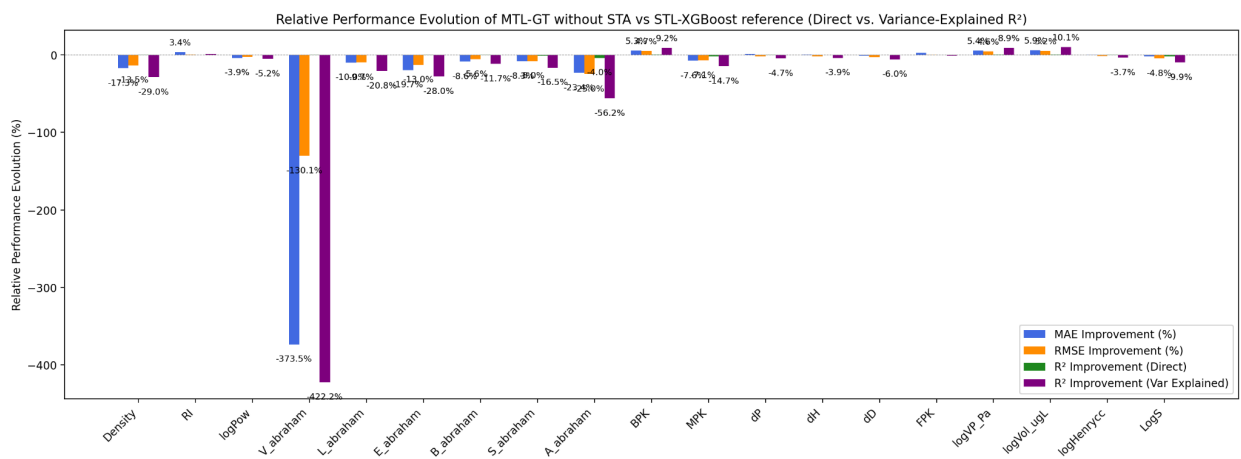

zero dotted line corresponds to the individual XGBoost result; evolution of performance between one given XGBoost and the graph transformer are reported in % from 3 metrics in barplot. To improve visibility on R^squared we proposed to look at the R^squared (Direct) and the 1-R^squared (Var Explained) effects.

## Discussion

As part of the features are highly correlated to our targets: Ahrabams. We observed that for almost all features XGBoost is equivalent to naive Graph Transformer without Synthetic Task Augmentation (see Figure 1). By Adding the Synthetic Task Augmentation, we observe an improvement on such tasks showing the capability of the method to leverage XGBoost knowledge into the Graph Transformer (see Figure 2).

We proposed a unified pipeline to train a model based on XGBoost knowledge coming from Osmordred features. Similar results were also observed for other types of features and other types of main trained model like RNN[24] unreported results.

While we anticipate that multitasks need to be applied to relative connected tasks we never really have a proof of the phenomena.

While V Abraham is still not as perfect as in XGBoost the Graph Transformer has a much better performance using the "synthetic target" showing that the model is able to transfer knowledge through the target task as expected.

Here, we systematically add a correlated task (synthetic pretrained) for each experiment task. By adding those synthetic tasks we see a high improvement of our model while more than 99.99% of the architecture is identical except the very last projection layer that needed to be adjusted to the number of tasks expected dimension so from 19 to 38.

Figure 2: comparison of Graph Transformer with or without Synthetic Task Augmentation

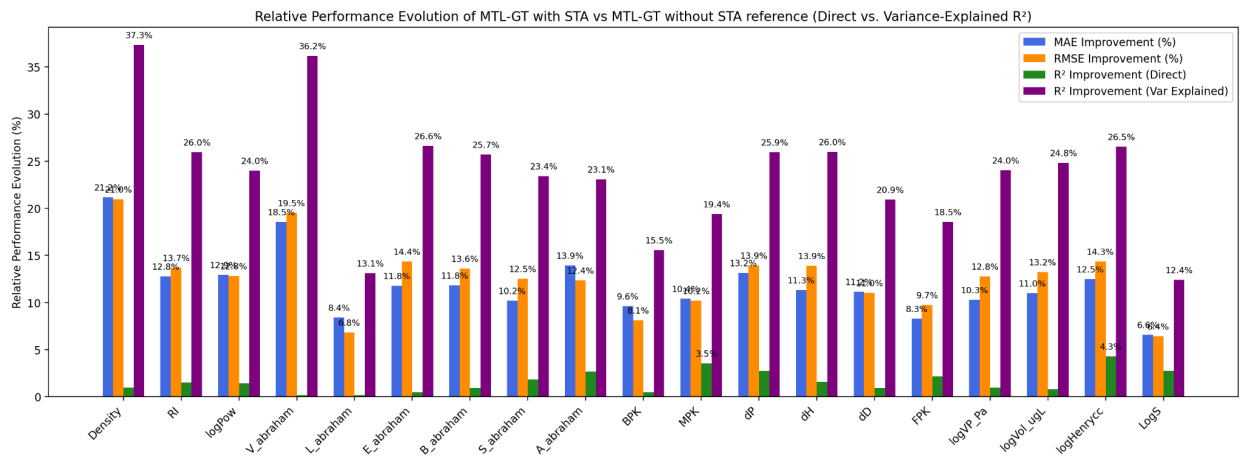

zero dotted line corresponds to one naive Graph Transformer without Synthetic Task Augmentation result; evolution of performance between zero dotted line reference and the Graph Transformer with Synthetic Task Augmentation are reported in % from 3 metrics in barplot. To improve visibility on R^squared we proposed to look at the R^squared (Direct) and the 1-R^squared (Var Explained) effects.

This method is robust and could be used without pretrained embedding or direct usage of the Osmordred features into the embedding space. It also demonstrates that we can transfer knowledge from the "XGBoost model" into the Graph Transformer model. This *in situ* synergistic effect is the first time demonstrated on a challenging multitask goal.

---

[24] Sepp Hochreiter and Jürgen Schmidhuber, "Long Short-Term Memory," *Neural Computation* 9, no. 8 (November 1, 1997): 1735–80, https://doi.org/10.1162/neco.1997.9.8.1735.

Interestingly, exact predictions of V and E Abraham features can be perfectly modeled by XGBoost (r2 > 0.99). With Synthetic Task Augmentation, we significantly reduce the gap to these near-perfect XGBoost results and observe a clear improvement over the naive Graph Transformer. Moreover, all inexact targets (i.e., features not included in the Osmordred set) show consistent performance gains (see Figure 3 in the Supplementary Information).

Synthetic task augmentation is by definition a data augmentation with a sparsity of 0, increasing the data applicability domain based on XGBoost knowledge, which should improve the model generalizability.

One potential explanation could be that synthetic task augmentation acting as a tutor/guide which may be seen as an approximation of a physical derivative equation PINNs[25] without the need to explicitly write such equations that are not fully known in general especially on so many tasks.

## Conclusion

We have demonstrated the synergetic effect of coupling synthetic and experimental tasks to train a single graph transformer to obtain excellent performance systematically on all tasks compared to naive graph transformers without synthetic task augmentation. Facilitating the usage of multitask single model and avoiding pretrain of embeddings or feature injection.

Intrinsically, the single task is a less complex task for a model. Based on two other collaborations that will be published soon applying synthetic task augmentation on both RNN or Transformer[26] architectures for single tasking, we obtained a boosting performance effect making synthetic task augmentation also applicable for single tasks and other architectures.

Moreover, we have demonstrated that on targets not highly correlated to features, XGBoost model can be surpassed by our Graph Transformer coupled to Synthetic task augmentation methodology that is not easy in machine learning.

---

[25] Maziar Raissi, Paris Perdikaris, and George Em Karniadakis, "Physics Informed Deep Learning (Part I): Data-Driven Solutions of Nonlinear Partial Differential Equations" (arXiv, 2017), https://doi.org/10.48550/ARXIV.1711.10561.

[26] Ashish Vaswani et al., "Attention Is All You Need," in *Advances in Neural Information Processing Systems*, ed. I. Guyon et al., vol. 30 (Curran Associates, Inc., 2017), https://proceedings.neurips.cc/paper_files/paper/2017/file/3f5ee243547dee91fbd053c1c4a845aa-Paper.pdf.

# Data Availability

As a preprint, data cannot be shared yet, the model will be shared upon demand to the author after publication.

# Funding

Author was funded by Osmo Labs PBC for Osmordred development and new models, architectures and methodology development research on machine learning.

# Competing Interests and Consent for publication

The author declares that he has no competing interests. Author has read and agreed to the published version of the manuscript.

# Other

The author partially presented this work based on a single cross validation during the CECAM AiChemist conference at Lausanne 29.4.2025.

# Supplementary Information:

We use the "5 x 5 CV" method proposed by Pat Walter, by applying a seed 5 CV split using random pandas settings for 5 seeds [3,5,7,13,42]. XGBoost was trained with those settings. For each seed, we average the 5 models to predict the 19 targets and use those ensemble target predictions as Synthetic Task Augmentation. For the Graph Transformer model we use an Early Stopping procedure limit of 20 and 450 max epochs using Adams optimizer and batch size of 64.
Tables are reflecting a "5 x CV 5" split variation for XGBoost (Table 4), naive Graph Transformer without Synthetic Task Augmentation (Table 5) and Graph Transformer with Synthetic Task Augmentation (Table 6). For MAE and RMSE lower the better for R2 closer to 1 the better. We report also the comparison of XGBoost and Graph Transformer Synthetic Task Augmentation in figure 3.

Table 4 : XGBoost performances (5 x 5 CV)

| Target | mae_mean | mae_std | rmse_mean | rmse_std | r2_mean | r2_std |
|---:|---:|---:|---:|---:|---:|---:|
| Density | 0.0173 | 0.0002 | 0.0416 | 0.0010 | 0.9807 | 0.0014 |
| RI | 0.0089 | 0.0002 | 0.0213 | 0.0004 | 0.9445 | 0.0023 |
| logPow | 0.2993 | 0.0035 | 0.4466 | 0.0146 | 0.9461 | 0.0031 |
| **V_abraham** | 0.0049 | 0.0002 | 0.0176 | 0.0018 | 0.9991 | 0.0002 |
| L_abraham | 0.1976 | 0.0035 | 0.3592 | 0.0184 | 0.9880 | 0.0013 |
| **E_abraham** | 0.0483 | 0.0007 | 0.0900 | 0.0025 | 0.9868 | 0.0006 |
| B_abraham | 0.0546 | 0.0012 | 0.0969 | 0.0021 | 0.9683 | 0.0011 |
| S_abraham | 0.1022 | 0.0010 | 0.1838 | 0.0025 | 0.9377 | 0.0016 |
| A_abraham | 0.0337 | 0.0006 | 0.0860 | 0.0032 | 0.9335 | 0.0053 |
| BPK | 10.0739 | 0.1849 | 20.4647 | 0.8453 | 0.9674 | 0.0023 |
| MPK | 26.3907 | 0.1178 | 37.2538 | 0.3485 | 0.8656 | 0.0035 |
| dP | 0.6184 | 0.0074 | 1.1690 | 0.0362 | 0.9087 | 0.0057 |
| dH | 0.5436 | 0.0064 | 1.0327 | 0.0336 | 0.9441 | 0.0040 |
| dD | 0.2074 | 0.0020 | 0.3599 | 0.0092 | 0.9603 | 0.0021 |
| FPK | 13.3650 | 0.1127 | 22.6412 | 0.2590 | 0.8964 | 0.0031 |
| logVP_Pa | 0.3682 | 0.0066 | 0.6272 | 0.0124 | 0.9575 | 0.0022 |
| logVol_ugL | 0.3768 | 0.0092 | 0.6377 | 0.0155 | 0.9655 | 0.0022 |
| logHenrycc | 0.8434 | 0.0289 | 1.5709 | 0.0675 | 0.8660 | 0.0064 |
| LogS | 0.6110 | 0.0045 | 0.9133 | 0.0092 | 0.8347 | 0.0046 |

Performances based on 5 x CV 5 individual/independent XGBoost model per target.

Table 5 : naive Graph Transformer Model without Synthetic Task Augmentation (5 x 5 CV)

| Target | mae_mean | mae_std | rmse_mean | rmse_std | r2_mean | r2_std |
|---:|---:|---:|---:|---:|---:|---:|
| Density | 0.0203 | 0.0008 | 0.0472 | 0.0019 | 0.9751 | 0.0020 |
| RI | 0.0086 | 0.0001 | 0.0212 | 0.0007 | 0.9449 | 0.0030 |
| logPow | 0.3111 | 0.0053 | 0.4579 | 0.0187 | 0.9433 | 0.0040 |
| **V_abraham** | 0.0232 | 0.0013 | 0.0405 | 0.0013 | 0.9953 | 0.0003 |
| L_abraham | 0.2174 | 0.0073 | 0.3940 | 0.0335 | 0.9855 | 0.0024 |
| **E_abraham** | 0.0578 | 0.0010 | 0.1017 | 0.0019 | 0.9831 | 0.0004 |
| B_abraham | 0.0593 | 0.0015 | 0.1023 | 0.0023 | 0.9646 | 0.0016 |
| S_abraham | 0.1107 | 0.0016 | 0.1985 | 0.0019 | 0.9274 | 0.0010 |
| A_abraham | 0.0416 | 0.0006 | 0.1075 | 0.0025 | 0.8961 | 0.0054 |
| BPK | **9.5448** | 0.1991 | 19.4980 | 0.7840 | 0.9704 | 0.0024 |
| MPK | 28.3911 | 0.2325 | 39.9037 | 0.3315 | 0.8458 | 0.0038 |
| dP | 0.6121 | 0.0093 | 1.1965 | 0.0404 | 0.9044 | 0.0059 |
| dH | 0.5404 | 0.0121 | 1.0521 | 0.0408 | 0.9419 | 0.0049 |
| dD | 0.2098 | 0.0038 | 0.3706 | 0.0110 | 0.9579 | 0.0029 |
| FPK | **13.0568** | 0.1871 | 22.7489 | 0.3132 | 0.8954 | 0.0032 |
| logVP_Pa | **0.3482** | 0.0078 | 0.5982 | 0.0227 | 0.9613 | 0.0035 |
| logVol_ugL | **0.3546** | 0.0084 | 0.6048 | 0.0227 | 0.9690 | 0.0028 |
| logHenrycc | 0.8469 | 0.0208 | 1.5995 | 0.0657 | 0.8610 | 0.0069 |
| LogS | 0.6240 | 0.0037 | 0.9575 | 0.0034 | 0.8184 | 0.0037 |

In **Bold** improvement compared to XGBoost models, using one GT model instead of 19 independent XGBoost models.

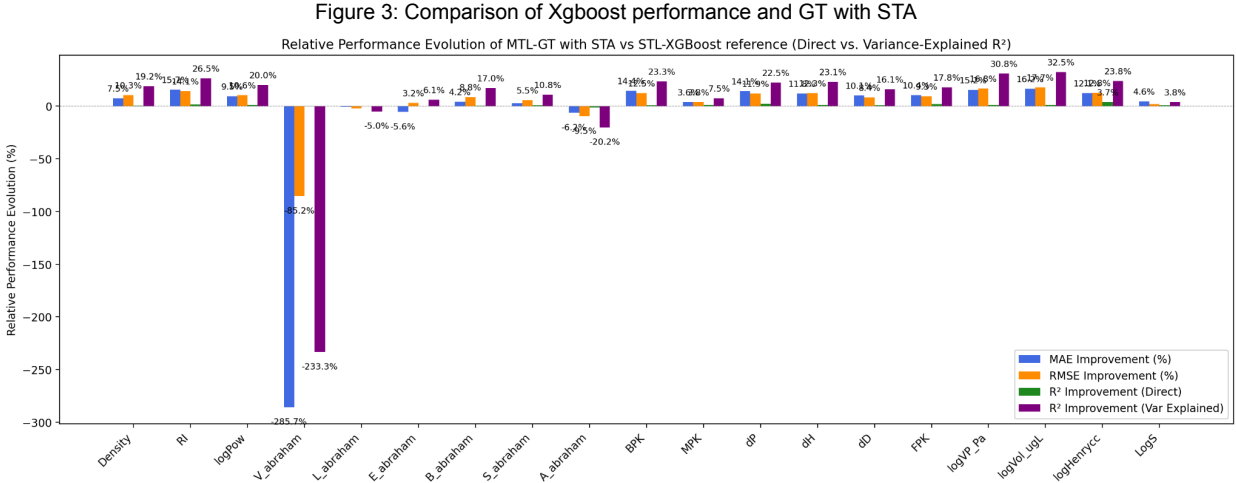

Figure 3: Comparison of Xgboost performance and GT with STA

Graph Transformer Synthetic Task Augmentation performs better and XGBoost on all targets except for several Abrahams features present in Osmordred.

Table 6 : Graph Transformer Model with Synthetic Task Augmentation  (5 x 5 CV)

| Target | mae_mean | mae_std | rmse_mean | rmse_std | r2_mean | r2_std |
|---|---|---|---|---|---|---|
| Density | **0.0160** | 0.0003 | 0.0373 | 0.0016 | 0.9844 | 0.0019 |
| RI | **0.0075** | 0.0001 | 0.0183 | 0.0005 | 0.9592 | 0.0025 |
| logPow | **0.2709** | 0.0050 | 0.3992 | 0.0166 | 0.9569 | 0.0032 |
| V_abraham | 0.0189 | 0.0009 | 0.0326 | 0.0009 | 0.9970 | 0.0001 |
| L_abraham | **0.1991** | 0.0045 | 0.3671 | 0.0433 | 0.9874 | 0.0029 |
| E_abraham | 0.0510 | 0.0010 | 0.0871 | 0.0017 | 0.9876 | 0.0004 |
| B_abraham | **0.0523** | 0.0008 | 0.0884 | 0.0009 | 0.9737 | 0.0004 |
| S_abraham | **0.0994** | 0.0023 | 0.1736 | 0.0026 | 0.9444 | 0.0012 |
| A_abraham | 0.0358 | 0.0008 | 0.0942 | 0.0049 | 0.9201 | 0.0087 |
| BPK | **8.6257** | 0.1571 | 17.9162 | 0.7901 | 0.9750 | 0.0020 |
| MPK | **25.4404** | 0.2939 | 35.8295 | 0.3360 | 0.8757 | 0.0032 |
| dP | **0.5315** | 0.0076 | 1.0299 | 0.0311 | 0.9292 | 0.0040 |
| dH | **0.4792** | 0.0068 | 0.9060 | 0.0321 | 0.9570 | 0.0034 |
| dD | **0.1864** | 0.0029 | 0.3297 | 0.0082 | 0.9667 | 0.0018 |
| FPK | **11.9723** | 0.1496 | 20.5322 | 0.3270 | 0.9148 | 0.0029 |
| logVP_Pa | **0.3124** | 0.0025 | 0.5218 | 0.0069 | 0.9706 | 0.0013 |
| logVol_ugL | **0.3157** | 0.0023 | 0.5248 | 0.0069 | 0.9767 | 0.0010 |
| logHenrycc | **0.7411** | 0.0206 | 1.3700 | 0.0625 | 0.8979 | 0.0073 |
| LogS | **0.5828** | 0.0039 | 0.8960 | 0.0081 | 0.8409 | 0.0053 |

Effect of Adding Synthetic Task Augmentation on GT model (green) improvement compare the GT without STA and in bold improvement compare to XGBoost model using Osmordred features.